\documentclass[conference,compsoc]{IEEEtran}

\ifCLASSINFOpdf
\else
\fi

\usepackage{amsmath}
\usepackage{graphicx} 
\usepackage{multirow}
\usepackage{algorithm}
\usepackage{algpseudocode}
\usepackage{url}
\usepackage[compress]{cite}
\usepackage{amsmath}
\usepackage{amsfonts}
\usepackage{graphicx} 
\usepackage{multirow}
\usepackage{algorithm}
\usepackage{algpseudocode}
\usepackage{balance}
\begin{document}

% ======== TITLE ========
\title{FastCentNN: Accelerating Centroid Neural Network with Entropy Proxy}

% Author
\author{Le-Anh Tran\\
\tt\small tranlevision@gmail.com
}

% make the title area
\maketitle

% ======== ABSTRACT ========
\begin{abstract}
Centroid neural network (CentNN) is an unsupervised competitive learning algorithm in which centroid splitting is triggered only after strict local stabilization, often leading to prolonged low-movement training phases before model expansion. This report proposes FastCentNN, an accelerated variant that addresses this inefficiency by introducing an early splitting strategy based on the total centroid movement per epoch, which serves as a training entropy proxy. As a result, FastCentNN reduces unnecessary reassignment epochs while preserving the original winner-loser learning dynamics. FastCentNN supports both absolute and stage-relative movement thresholds, allowing the splitting criterion to remain either fixed or adaptive throughout training. Experiments on some benchmark datasets show that FastCentNN consistently achieves clustering quality comparable to CentNN while reducing runtime by up to 16\% on synthetic 2D datasets and about 5\% on high-dimensional datasets. FastCentNN therefore provides a practical and efficient drop-in replacement for CentNN, retaining its online adaptive learning behavior while offering a simple and interpretable speed-stability trade-off through configurable splitting thresholds. Code is available\footnote{\url{https://github.com/tranleanh/centroid-neural-networks}}.
\end{abstract}

\IEEEpeerreviewmaketitle

% ======== BODY ========
% ===================================================================================
\section{Introduction}
\label{sec:intro}

Clustering remains a core task in unsupervised learning, with applications spanning pattern discovery, compression, anomaly detection, and post-processing 
\cite{macqueen1967some,tran2019robust,park2000centroid,Tran2023,tran2024clustering,tran2024cluster}. 
Among modern clustering approaches, centroid neural network (CentNN) \cite{park2000centroid} is a prototype-based algorithm that follows a competitive-learning paradigm, where centroid vectors are updated online through winner-loser dynamics. This design is well suited to incremental data processing, memory-constrained environments, and adaptive model growth, placing CentNN within the broader family of competitive and self-organizing neural models \cite{kohonen1990self,martinetz1993neural}. Unlike batch clustering methods, CentNN updates centroids one sample at a time and expands model capacity through centroid splitting.

\begin{figure*}
\centering
\includegraphics[width=0.85\textwidth]{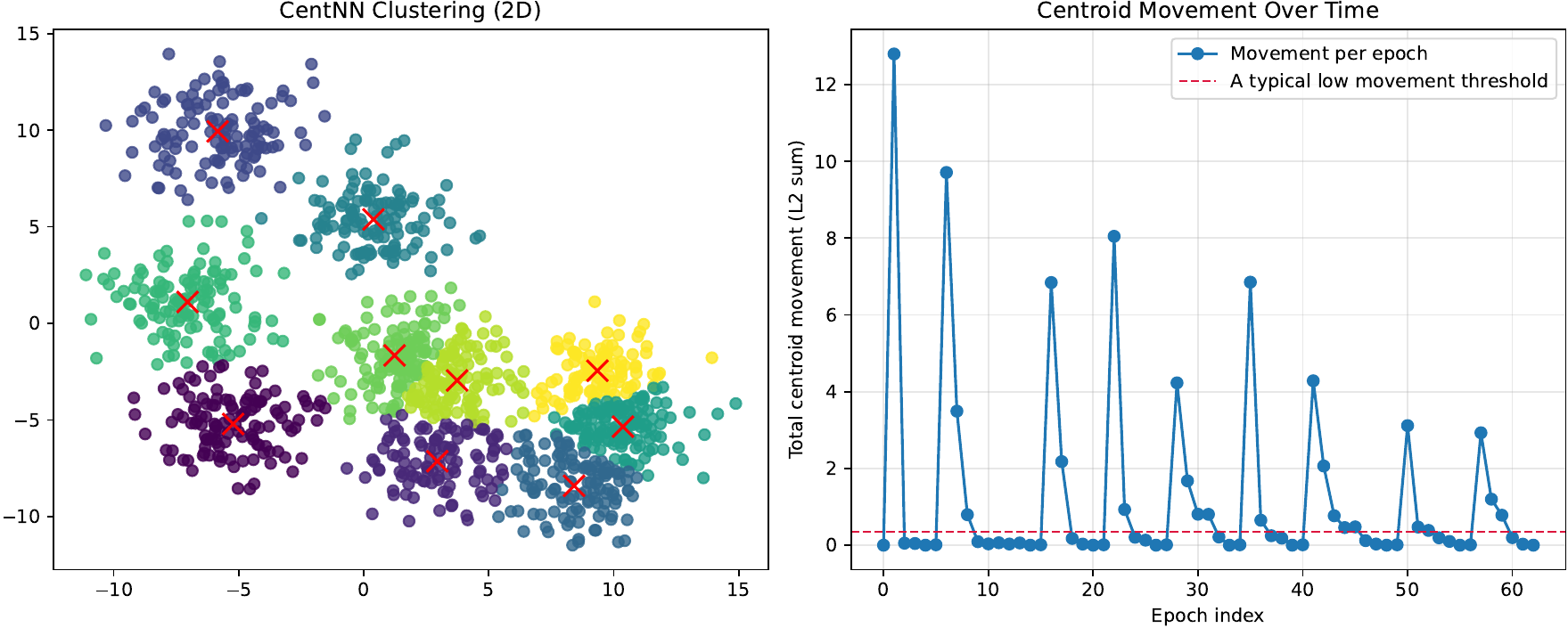}
\caption{Clustering result (left) and convergence behavior (right) of CentNN on a synthetic 2D dataset. CentNN converges through occasional large centroid adjustments after split events, separated by periods of minimal movement.}
\label{fig:centnn_convergence}
\end{figure*}

Despite these advantages, standard CentNN suffers from a practical inefficiency: before each split event, it may spend many epochs in a low-activity regime where centroid movement is minimal, yet splitting is delayed until strict local stabilization, as illustrated in Fig.~\ref{fig:centnn_convergence}. These ``flat'' phases increase runtime without corresponding structural progress. To address this limitation, this report proposes FastCentNN, a more efficient variant that treats global centroid movement per epoch as an entropy proxy for clustering uncertainty and initiates splitting once movement remains below a threshold for a patience window. FastCentNN converts split timing from a strict convergence event into an adaptive early-trigger decision while preserving the original winner-loser learning dynamics. It supports both absolute and stage-relative thresholds, enabling consistent behavior across datasets and training stages. Although the asymptotic computational complexity remains unchanged, FastCentNN improves practical efficiency by reducing unnecessary low-information epochs between split operations.

% \begin{algorithm*}[ht]
% \caption{Fast Centroid Neural Network (FastCentNN)}
% \label{alg:fast-centnn}
% \begin{algorithmic}[1]
% \Require Dataset $\mathcal{X}$, target number of clusters $K$, max epochs $T$, split scale $\epsilon$, movement threshold $\tau$, patience $P$
% \State Initialize two centroids near the global mean
% \State Initialize cluster assignments and baseline movement $B$
% \For{$t=1$ to $T$}
%     \State Perform one epoch of winner-loser centroid updates
%     \State Compute assignment-change count $C_t$
%     \State Compute total centroid movement $M_t$
%     \State Update patience counter using $M_t$ and threshold $\tau_t$
%     \State $\texttt{splitTrigger} \leftarrow (\texttt{patience} \ge P)\ \lor\ (C_t = 0)$
%     \If{$k_t < K$ \textbf{and} \texttt{splitTrigger}}
%         \State Select the cluster with the largest within-cluster error
%         \State Split the selected centroid using perturbation scale $\epsilon$
%         \State Reset baseline movement $B$ and patience counter
%     \ElsIf{$C_t = 0$}
%         \State \textbf{break}
%     \EndIf
% \EndFor
% \State \Return final centroids and cluster assignments
% \end{algorithmic}
% \end{algorithm*}

% ===================================================================================
\section{FastCentNN}
\label{sec:method}

\subsection{Problem Setup}
Let $\mathcal{X}=\{x_i\}_{i=1}^n$, with $x_i\in\mathbb{R}^d$, and let $K$ be the target number of clusters.  
CentNN represents clusters by centroids $\{w_j\}_{j=1}^{k_t}$ at epoch $t$, where $k_t$ grows over time via split operations. For each sample $x$, the winner index is defined as:
\begin{equation}
j^\star(x)=\arg\min_{j\in\{1,\dots,k_t\}} \|x-w_j\|_2.
\end{equation}
Each sample also stores its previous assignment $a_i^{(t-1)}$ from the previous epoch.

\subsection{Winner-Loser Updates}
If sample $x_i$ changes assignment at epoch $t$ from previous cluster $p=a_i^{(t-1)}$ to current winner $c=j^\star(x_i)$, the centroids are updated by:
\begin{align}
w_c &\leftarrow w_c + \frac{1}{N_c+1}(x_i-w_c), \\
w_p &\leftarrow w_p - \frac{1}{N_p-1}(x_i-w_p),
\end{align}
where $N_c,N_p$ are current member counts before reassignment.  
The loser update is applied only when $N_p>1$ to avoid division by zero and unstable singleton collapse.

\subsection{Split Operation}
CentNN starts from two centroids near the global mean:
\begin{equation}
w_1=\bar{x}+\epsilon,\quad w_2=\bar{x}-\epsilon,\quad
\bar{x}=\frac{1}{n}\sum_{i=1}^n x_i,
\end{equation}
where $\epsilon$ is a small perturbation. The centroids are updated iteratively until local stabilization is reached. If $k_t<K$, the split candidate is selected based on the maximum within-cluster error:
\begin{equation}
s=\arg\max_j E_j,\quad
E_j=\sum_{x_i:a_i=j}\|x_i-w_j\|_2.
\end{equation}
Then a symmetric split is applied:
\begin{equation}
w_s^- = w_s-\delta_s,\qquad
w_s^+ = w_s+\delta_s,
\end{equation}
where $\delta_s$ is a small perturbation (magnitude controlled by $\epsilon$), typically aligned with the largest-variance feature within cluster $s$.

\begin{table*}[t]
\centering
\caption{Performance of \{CentNN / FastCentNN\} on synthetic 2D datasets.}
\label{tab:synthetic_results}
\begin{tabular}{ccccccc}
\hline
Dataset & Clusters & Instances & Runtime (s) & Epochs & $\Delta$MSE (\%) & Speed Up \\
\hline
A1    & 20 & 3,000 & 1.575 / \textbf{1.312} & 155 / \textbf{128} & +2.47$\times10^{-7}$ & 16.72\% \\
A2    & 35 & 5,250 & 4.771 / \textbf{4.324} & 274 / \textbf{246} & +5.18$\times10^{-7}$ & 9.37\% \\
S1    & 15 & 5,000 & 1.110 / \textbf{1.048} & 64 / \textbf{61} & 0 & 5.54\% \\
S2    & 15 & 5,000 & 1.856 / \textbf{1.728} & 110 / \textbf{102} & +7.53$\times10^{-7}$ & 6.90\% \\
R15   & 15 & 600   & 0.077 / \textbf{0.073} & 34 / \textbf{33} & 0 & 4.18\% \\
Aggregation & 7  & 788   & 0.113 / \textbf{0.096} & 40 / \textbf{34} & -2.83$\times10^{-2}$ & 14.80\% \\
\hline
\end{tabular}
\end{table*}

\begin{table*}[t]
\centering
\caption{Performance of \{CentNN / FastCentNN\} on MNIST and Fashion-MNIST.}
\label{tab:mnist_fashion_results}
\begin{tabular}{ccccccc}
\hline
Dataset & Clusters & Instances & Runtime (s) & Epochs & $\Delta$MSE (\%) & Speed Up \\
\hline
MNIST & 10 & 10,000 & 8.344 / \textbf{7.924} & 141 / \textbf{132} & +1.53$\times10^{-5}$ & 5.03\% \\
Fashion-MNIST & 10 & 10,000 & 8.220 / \textbf{7.775} & 127 / \textbf{119} & -6.37$\times10^{-5}$ & 5.41\% \\
\hline
\end{tabular}
\end{table*}

\subsection{Early Split in FastCentNN}
Standard CentNN often spends many epochs in a low-activity regime with minimal centroid movement before satisfying the splitting criterion. FastCentNN addresses this inefficiency by replacing strict stabilization with an early splitting criterion based on the total centroid movement:
\begin{equation}
M_t=\sum_{j=1}^{k_t}\|w_j^{(t)}-w_j^{(t-1)}\|_2,
\end{equation}
where $M_t$ is the global centroid movement (entropy proxy) at epoch $t$. A split is triggered when $M_t$ remains below a threshold for a predefined patience window:
\begin{equation}
M_t \le \tau_t \text{ for } P \text{ consecutive epochs},
\end{equation}
where $\tau_t$ is the movement threshold and $P$ is the patience parameter. The threshold can be configured in two modes. In the \textit{absolute} mode, $\tau_t=\tau$, whereas in the \textit{stage-relative} mode, $\tau_t=\tau B_r$, where $B_r$ is the baseline movement measured immediately after the latest split stage $r$. Following each split, the stage state is reset as:
\begin{equation}
B_{r+1}\leftarrow \max(M_t,\varepsilon_0),\qquad
\text{streak}\leftarrow 0,
\end{equation}
where $\varepsilon_0>0$ is a small constant for numerical stability. The full FastCentNN procedure is summarized in Algorithm~\ref{alg:fast-centnn}.

\begin{algorithm}[t]
\caption{Fast Centroid Neural Network (FastCentNN)}
\label{alg:fast-centnn}
\begin{algorithmic}[1]
\Require Dataset $\mathcal{X}$, target num of clusters $K$, max epochs $T$, split scale $\epsilon$, movement threshold $\tau$, patience $P$
\State Initialize two centroids near the global mean
\State Initialize cluster assignments and baseline movement $B$
\For{$t=1$ to $T$}
    \State Perform one epoch of winner-loser centroid updates
    \State Compute assignment-change count $C_t$
    \State Compute total centroid movement $M_t$
    \State Update patience counter using $M_t$ and $\tau_t$
    \State $\texttt{split} \leftarrow (\texttt{patience} \ge P)\ \lor\ (C_t = 0)$
    \If{$k_t < K$ \textbf{and} \texttt{split}}
        \State Pick the cluster with max within-cluster error
        \State Split the selected cluster's centroid using $\epsilon$
        \State Reset $B$ and patience counter
    \ElsIf{$C_t = 0$}
        \State \textbf{break}
    \EndIf
\EndFor
\State \Return final centroids and cluster assignments
\end{algorithmic}
\end{algorithm}

% ===================================================================================
\section{Experiments}
\label{sec:experiments}

\subsection{Experimental Setup}

FastCentNN is evaluated against the original CentNN under identical parameter settings on two clustering tasks: (i) synthetic 2D datasets and (ii) high-dimensional datasets. Since the primary objective is to improve computational efficiency while preserving clustering quality, the relative difference in mean squared error (MSE) is reported as:
\begin{equation}
\Delta\mathrm{MSE}(\%) =
\frac{\mathrm{MSE}_{\text{FastCentNN}} - \mathrm{MSE}_{\text{CentNN}}}
{\mathrm{MSE}_{\text{CentNN}}}
\times 100.
\label{eq:delta_mse}
\end{equation}
In addition, efficiency is assessed using the total runtime and the number of training epochs required for convergence.

\subsection{Test on Synthetic 2D Data}

\begin{figure*}
\centering
\includegraphics[width=\textwidth]{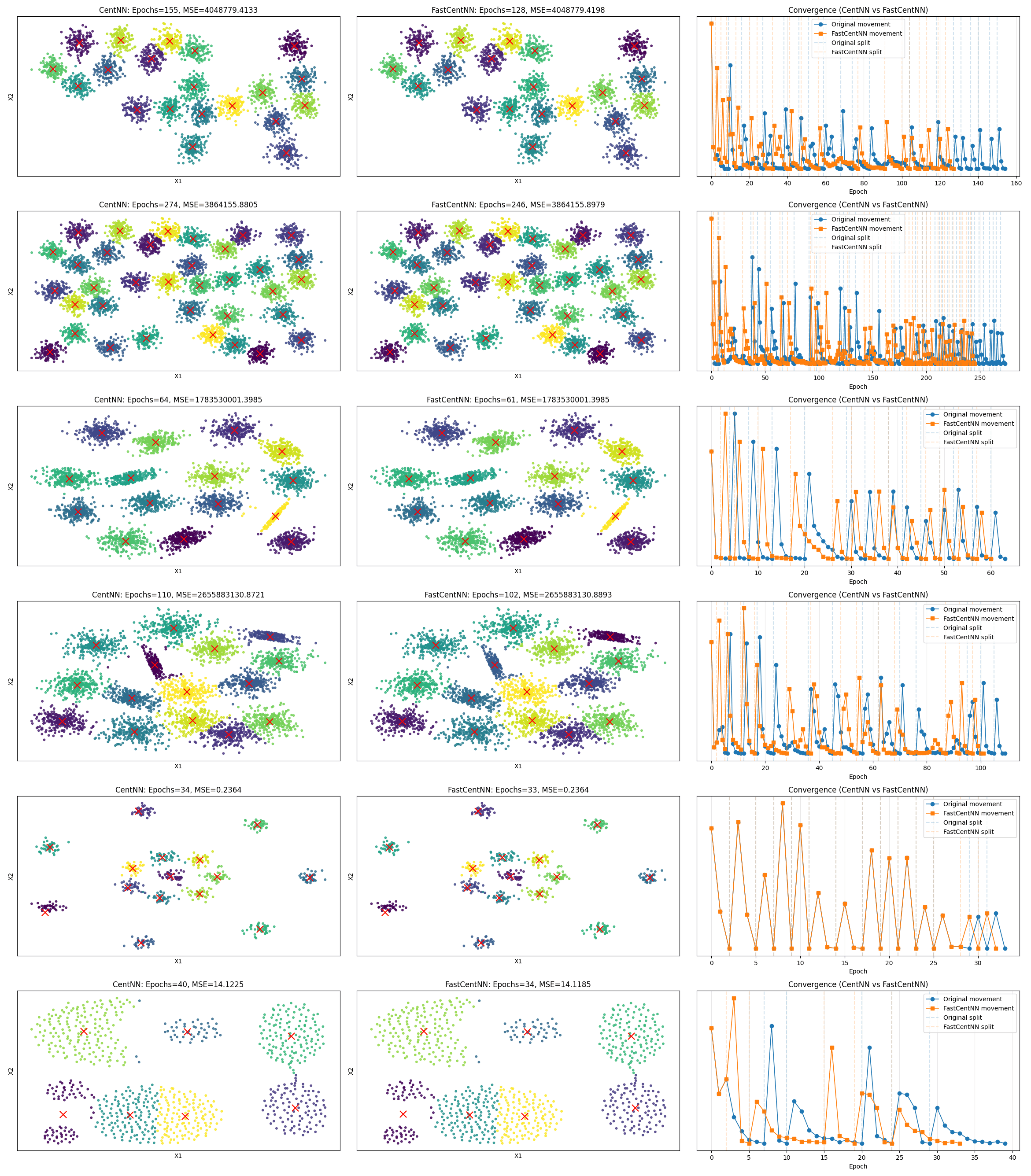}
\caption{Comparison on synthetic 2D datasets.}
\label{fig:comparison_2D}
\end{figure*}

FastCentNN was evaluated against the original CentNN on six widely used synthetic clustering benchmark datasets: A1, A2, S1, S2, R15, and Aggregation \cite{ClusteringDatasets}. Each dataset consists of 2D data points with a predefined number of clusters. Table~\ref{tab:synthetic_results} summarizes the results. Compared with the original CentNN, FastCentNN consistently reduces both runtime and the number of training epochs while producing nearly identical clustering quality. Across all six datasets, FastCentNN achieves an average runtime reduction of approximately $9.6\%$, with the largest speed-up of $16.72\%$ on A1 and the smallest of $4.18\%$ on R15. The $\Delta\mathrm{MSE}$ measures are negligible for all datasets, remaining close to zero even on the most challenging cases. These results indicate that the proposed early splitting strategy effectively eliminates unnecessary low-movement epochs, improving computational efficiency without degrading clustering performance. Visual results are shown in Fig. \ref{fig:comparison_2D}.

\subsection{Test on High-dimensional Data}

FastCentNN was also evaluated on two high-dimensional image datasets, MNIST and Fashion-MNIST, each configured with 10 target clusters. Table~\ref{tab:mnist_fashion_results} summarizes the results. FastCentNN consistently reduces both runtime and the number of training epochs while preserving clustering quality. On MNIST, the runtime is improved from 8.344\,s to 7.924\,s, corresponding to a speed-up of 5.03\%, while the number of training epochs is decreased from 141 to 132. Similarly, on Fashion-MNIST, FastCentNN reduces the runtime from 8.220\,s to 7.775\,s (5.41\% speed-up) and the number of training epochs from 127 to 119.

The $\Delta\mathrm{MSE}$ values are negligible on both datasets, remaining within $10^{-5}\%$ compared to the original CentNN. These results demonstrate that the proposed early splitting strategy remains effective on high-dimensional datasets, reducing computational cost while maintaining overall clustering accuracy.

% ===================================================================================
\section{Conclusion}
\label{sec:conclusion}

This report proposes FastCentNN, an accelerated variant of CentNN that introduces an early splitting strategy based on total centroid movement. By triggering centroid splits before strict local stabilization, FastCentNN reduces unnecessary low-movement epochs while preserving the original winner-loser learning dynamics. Experimental results on synthetic 2D datasets, MNIST, and Fashion-MNIST demonstrate consistent reductions in runtime and training epochs with negligible differences in clustering quality.

\section*{Acknowledgment}
The author would like to thank Prof. Dong-Chul Park, the inventor of the original CentNN algorithm, whose insights inspired the development of this approach.

\balance

% References
\bibliographystyle{IEEEtran}
% \footnotesize
\bibliography{ref.bib}
% \bibliography{ref}

% that's all folks
\end{document}